\documentclass[11pt,a4paper,fleqn]{article}
\usepackage[hyperref]{acl2020}
\usepackage{times}
\usepackage{latexsym}
\usepackage{linguex}
\usepackage{graphicx, graphbox}
\usepackage{tikz-qtree} % do not center trees
\usepackage{tikz-dependency}
\usepackage{booktabs}
\usepackage{subcaption}
\usepackage{amsmath}
\DeclareMathOperator*{\argmin}{arg\,min}

\usepackage{microtype}

\aclfinalcopy % Uncomment this line for the final submission
\title{Representations of Syntax [MASK] Useful:\\Effects of Constituency and Dependency Structure in Recursive LSTMs}

\author{Michael A. Lepori$^1$\qquad Tal Linzen$^{1,2}$ \qquad  R. Thomas McCoy$^2$ \\
  $^1$Department of Computer Science\qquad $^2$Department of Cognitive Science\\
  Johns Hopkins University\\
  \texttt{\{mlepori1, tal.linzen, tom.mccoy\}@jhu.edu}}

\date{}

\begin{document}
\maketitle
\begin{abstract}
Sequence-based neural networks show significant sensitivity to syntactic structure, but they still perform less well on syntactic tasks than tree-based networks. 
Such tree-based networks can be provided with a constituency parse, a dependency parse, or both. 
We evaluate which of these two representational schemes more effectively introduces biases for syntactic structure that increase performance on the subject-verb agreement prediction task.
We find that a constituency-based network generalizes more robustly than a dependency-based one, and that combining the two types of structure does not yield further improvement. 
Finally, we show that the syntactic robustness of sequential models can be substantially improved by fine-tuning on a small amount of constructed data, suggesting that data augmentation is a viable alternative to explicit constituency structure for imparting the syntactic biases that sequential models are lacking.

\end{abstract}

\setlength{\Exlabelwidth}{0.5em}
\setlength{\SubExleftmargin}{1.35em}

\section{Introduction }
\label{sec:intro}

Natural language syntax is structured hierarchically, rather than sequentially \cite{syntstructsChomsky57a, everaert2015structures}. One phenomenon that illustrates this fact is \textbf{English subject-verb agreement}, the requirement that verbs and their subjects must match in number. The hierarchical structure of a sentence determines which noun phrase each verb must agree with; sequential heuristics such as agreeing with the most recent noun may succeed on simple sentences such as~\ref{sent1} but fail in more complex cases such as~\ref{sent2}:

\ex. \a. The \textbf{boys kick} the ball. \label{sent1}
\b. The \textbf{boys} by the red truck \textbf{kick} the ball. \label{sent2} 

\noindent
We investigate whether a neural network must process input according to the structure of a syntactic parse in order for it to learn the appropriate rules governing these dependencies, or whether there is sufficient signal in natural language corpora for low-bias networks (such as sequential LSTMs) to learn these structures. We compare sequential LSTMs, which process sentences from left to right, with tree-based LSTMs that process sentences in accordance with an externally-provided, ground-truth syntactic structure. 

We consider two types of syntactic structure: \textbf{constituency structure} \cite{chomsky1993minimalist, pollard1994head} and \textbf{dependency structure} \cite{tesniere1959elements, hudson1984word}. We investigate models provided with either structure, both structures, or neither structure (see Table~\ref{table:twobytwo}), and assess how robustly these models learn subject-verb agreement when trained on natural language.\footnote{Code, data, and models are at \url{https://github.com/mlepori1/Representations_Of_Syntax}}

Even with the syntactic biases present in tree-based LSTMs, it is possible that natural language might not impart a strong enough signal to teach a network how to robustly track subject-verb dependencies. How might the performance of these tree-based LSTMs change if they were fine-tuned on a small dataset designed to impart a stronger syntactic signal? Furthermore, would we still need these tree structures, or could a sequential LSTM now learn to track syntactic dependencies?

\begin{table}[t]
\begin{center}
\resizebox{0.4\textwidth}{!}{
 \begin{tabular}{p {1cm}  p {2.7cm} p {2.8cm}} 
 \toprule
   & No Constituency & Constituency  \\ [0.5ex] 
 \midrule
  No Heads& BiLSTM & Constituency LSTM  \vspace{0.15in}\\ 
 Heads & Dependency LSTM & Head-Lexicalized LSTM  \\  [1ex] 
\bottomrule
\end{tabular}}
 \caption{Linguistic properties of our four models.}
\label{table:twobytwo} 
\end{center}
\end{table}

We find that building in either type of syntactic structure improves performance over the \mbox{BiLSTM} baseline, thus showing that these structures are learned imperfectly (at best) by low-bias models from natural language data. Of the two types of structure, constituency structure turns out to be more useful. The dependency-only model performs well on natural language test sets, but fails to generalize to an artificially-constructed challenge set.
After fine-tuning on a small dataset that is designed to impart a strong syntactic signal, the BiLSTM generalizes more robustly, but still falls short of the tree-based LSTMs.

We conclude that for a network to robustly show sensitivity to syntactic structure, stronger biases for syntactic structure need to be introduced than are present in a low-bias learner such as a BiLSTM, and that, at least for the subject-verb agreement task, constituency structure is more important than dependency structure. 
Both tree-based model structure and data augmentation appear to be viable approaches for imparting these biases.

\section{Related Work}
\label{sec:relatedwrk}
Prior work has shown that neural networks without explicit mechanisms for representing syntactic structure can show considerable sensitivity to syntactic dependencies \cite{goldberg2019assessing, gulordava-etal-2018-colorless, linzen-etal-2016-assessing}, and that certain aspects of the structure of the sentence can be reconstructed from their internal representations \cite{lin-etal-2019-open, giulianelli-etal-2018-hood, hewitt-manning-2019-structural}. \citet{marvin2018targeted} showed that sequential models still have substantial room for improvement in capturing syntax, and other work has shown that models with a greater degree of syntactic structure outperform sequential models on syntax-sensitive tasks \cite{yogatama2018memory, kuncoro-etal-2018-lstms, kuncoro-etal-2017-recurrent}, including some of the tree-based models used here \cite{Bowman:2015:TCN:2996831.2996836, li-etal-2015-tree}. One contribution of the present work is to tease apart the two major types of syntactic structure to see which one imparts more effective syntactic biases.

\section{Models}
\label{sec:models}
\subsection{BiLSTM}
\label{sec:lstmsection}
As our baseline model, we used a simple extension to the LSTM architecture \cite{Hochreiter:1997:LSM:1246443.1246450}, the \textbf{bidirectional LSTM} \cite[BiLSTM;][]{schuster97bidirectional}. This model runs one LSTM from left to right over a sequence, and another from right to left, without appealing to tree structure. Bidirectional LSTMs outperform unidirectional LSTMs on a variety of tasks \cite{huang2015bidirectional,chiu2016named}, including syntax-sensitive tasks \cite{kiperwasser2016simple}. \citet{ravfogel2019studying} also employs BiLSTMs for a similar agreement task. 

\subsection{Tree LSTMs}
\label{sec:treelstmsect}

To study the effects of explicitly building tree structure into the model architecture, we used the \textbf{Constituency LSTM} and the \textbf{Dependency LSTM} \cite{tai-etal-2015-improved}, which are types of recursive neural networks \cite{goller1996learning}.
The Constituency LSTM operates in accordance with a binary constituency parse, composing together vectors representing a left child and a right child into a vector representing their parent. Models similar to the Constituency LSTM have been proposed by \newcite{le-zuidema-2015-compositional} and \newcite{ Zhu:2015:LSM:3045118.3045289}. 

In a Dependency LSTM, the representations of a head's children are summed, and then composed with the representation of the head itself to yield a representation of the phrase that has that head.
See Appendix~\ref{sec:appEqs} for more details on both models.

\subsection{Head-Lexicalized Tree LSTMs}
\label{sec:headlstmsect}

To create a model where composition is simultaneously guided by both a dependency parse and a constituency parse, we modified the constituency model described in Section~\ref{sec:treelstmsect}, turning it into a  \textbf{head-lexicalized tree LSTM}. In a standard Constituency LSTM, the input for all non-leaf nodes is a vector of all 0's. To add head lexicalization, we instead feed in the word embedding of the correct headword of that constituent as the input, where the choice of headword is determined using the Stanford Dependency Parser \cite{manningparser}. See Appendix~\ref{sec:appHeadAlg} for more details, as well as an example of a head-lexicalized constituency tree. This model is similar to the head-lexicalized tree LSTM of \newcite{teng-zhang-2017-head}. However, their model learns how to select the heads of constituents in an unsupervised manner; these heads may not correspond to the syntactic notion of heads. Because we seek to understand the effect of using the heads derived from the dependency parse, we provide our models with explicit head information.

\section{\label{sec:exps}Task}
\label{sec:tasksec} We adapted a syntax-sensitive task that previous work has used to assess the syntactic capabilities of LSTMs---the number prediction task \cite{linzen-etal-2016-assessing}. The most standard version of this task is based on a left-to-right language modeling objective; however, tree-based models are not compatible with left-to-right language modeling. Therefore, we made two modifications to this objective, both of which have precedents in the literature: First, we gave the model an entire present-tense sentence with main verb masked out, following \citet{goldberg2019assessing}. Second, the model's target output was the number of the masked verb: \textsc{singular} or \textsc{plural}; we follow \citet{linzen-etal-2016-assessing} and \citet{ravfogel2019studying} in framing number prediction as a classification task. 
To solve the task, the model must identify the subject whose head is the main verb (in the dependency formalism), and use that information to determine the syntactic number of the verb; e.g., for~\ref{taskexa}, the answer is \textsc{singular}.
\ex. The girl \textbf{*MASK*} the ball. \label{taskexa}

\noindent
\newcite{linzen-etal-2016-assessing} pointed out that there are several incorrect heuristics which models might adopt for this task because these heuristics still produce decent classification accuracy. 
One salient example is picking the syntactic number of the most recent noun to the left of the verb. 
We hypothesize that tree-based models will be less susceptible to these non-robust heuristics than sequential models.

\section{Experiment 1: Natural Language}
\label{sec:exp1}

\paragraph{Data:}\label{sec:exp1data} We train our models on a subset of the dataset from \citet{linzen-etal-2016-assessing} that is chosen to have a uniform label distribution (50\% \textsc{singular} and 50\% \textsc{plural}). We made this choice because our task format differs from that used in some past work (see Section~\ref{sec:tasksec}), so performance on the task as we have framed it cannot be directly compared to prior work. 
In the absence of baselines from the literature, we use chance performance of 50\% as a baseline; to ensure that this baseline is reasonable, we balance the label distribution during training to discourage models from becoming biased toward one label.

We use two types of test sets: those that contain adversarial attractors, and those that do not. An \textbf{adversarial attractor} is a noun that is between the subject and the main verb of a sentence and that has the opposite syntactic number from the subject noun. Adversarial attractors have been found to produce agreement errors in humans \cite{Bock1991BrokenA} and neural models \cite{goldberg2019assessing, gulordava-etal-2018-colorless, linzen-etal-2016-assessing}. 
We use code from \newcite{goldberg2019assessing}\footnote{\url{https://github.com/yoavg/bert-syntax}} to extract adversarial datasets containing varying numbers of attractors, from~0 to~4 attractors.
Sentence~\ref{attr1} provides an example of a sentence with 4 attractors. 
\ex. Algorithmic \textbf{problems} such as \textit{[type]} \textit{[checking]} and \textit{[type]} \textit{[inference]} \textbf{are} more difficult for equirecursive types as well. \label{attr1}

See Appendix~\ref{app:data} for details on our corpus and on preprocessing, and Appendix~\ref{app:exp1train} for training.

\begin{figure}
    \begin{subfigure}{\columnwidth}
    \centering
    \includegraphics[width=4.5cm, align=t]{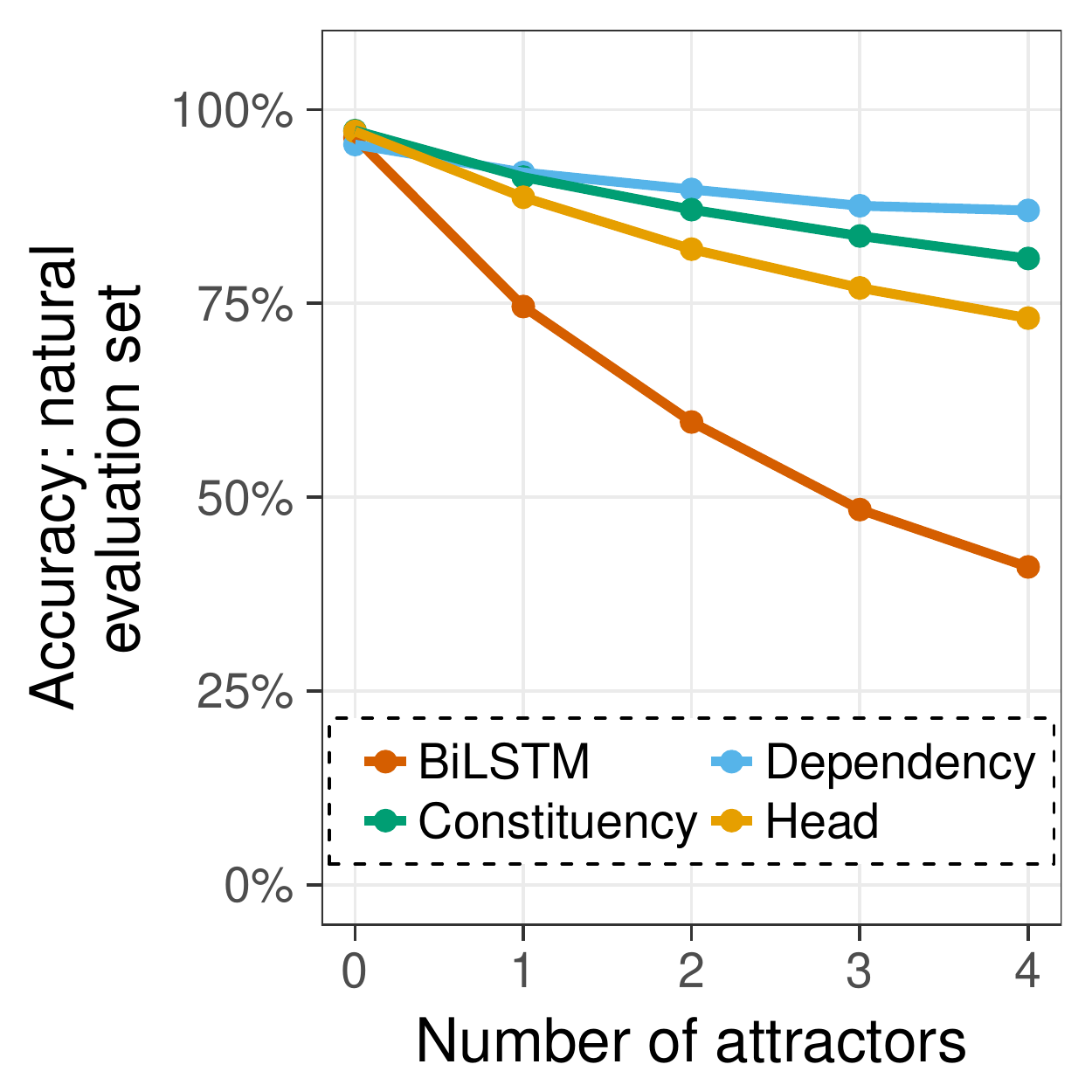}
    \hfill
    \includegraphics[width=2.7cm, height=5.11cm ,align=t]{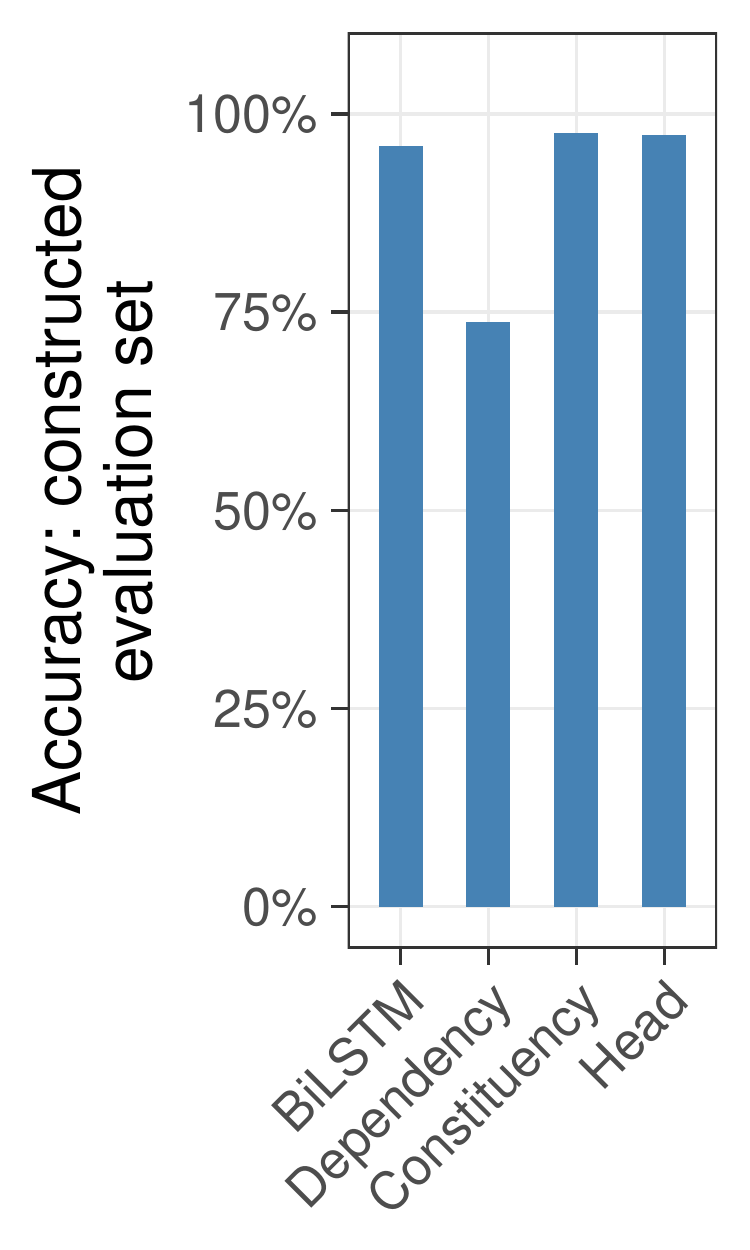}
    \caption{Results for models trained on natural language.}
    \label{fig:natFig}
    \end{subfigure}%
    \hfill
    \begin{subfigure}{\columnwidth}
    \centering
    \includegraphics[width=4.5cm, align=t]{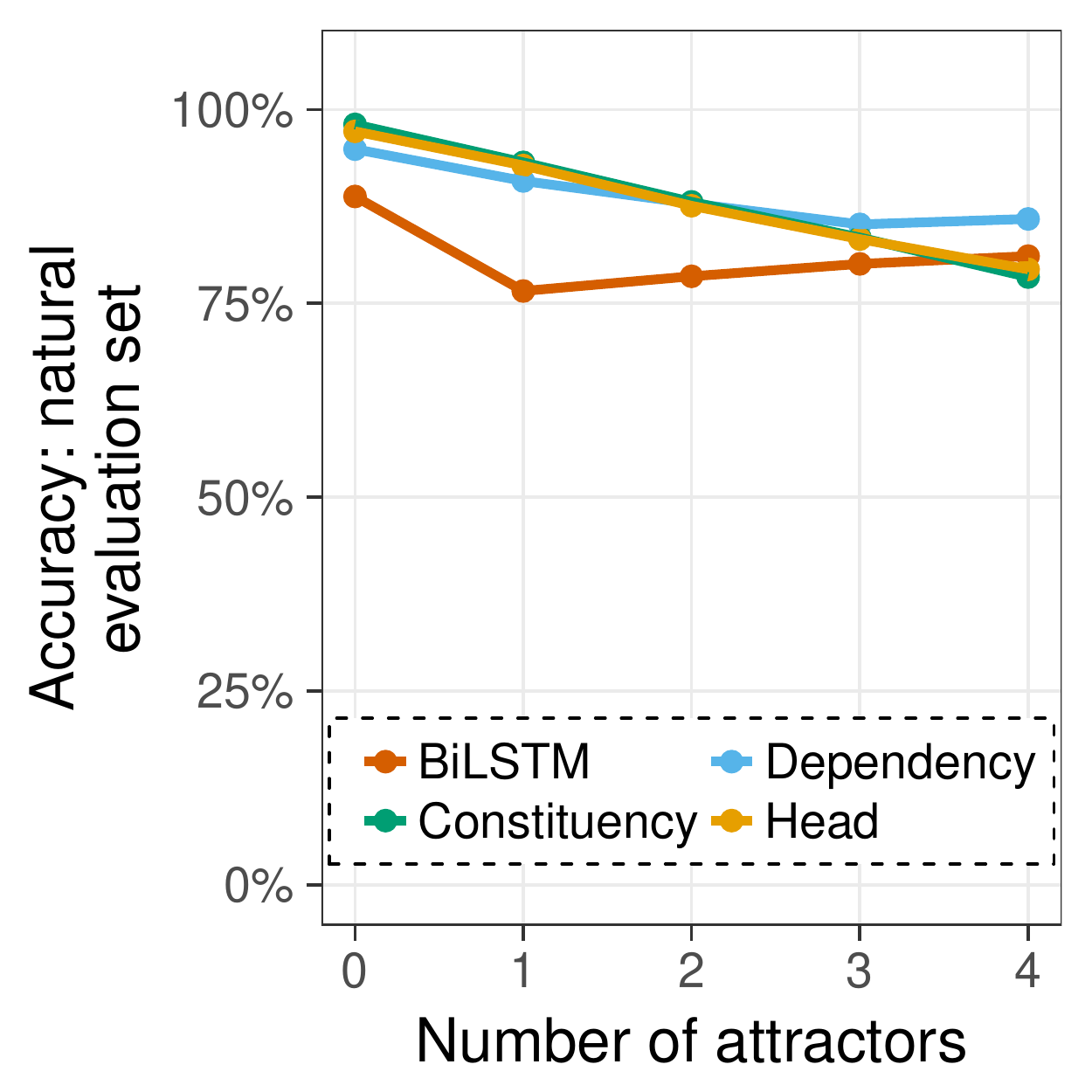}
    \hfill
    \includegraphics[width=2.7cm, height=5.11cm, align=t]{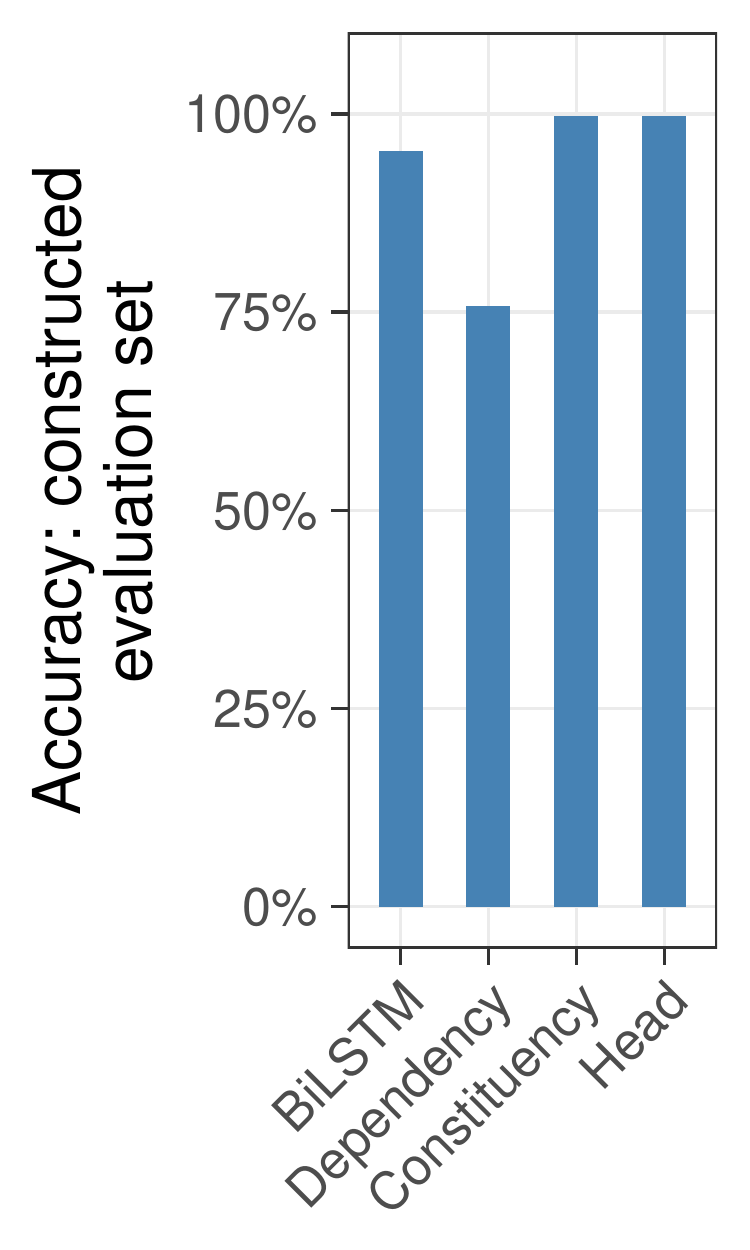}
    \caption{Results for models trained on natural language and then exposed to a 500-sentence augmentation set.}
    \label{fig:fineFig}
    \end{subfigure}
    \caption{Results on binary classification of masked verbs as \textsc{singular} or \textsc{plural}. All results are averages across 3 runs. Chance performance is 50\%.}
    \label{fig:resFig}
\end{figure}

\paragraph{Natural language evaluation:} All of the tree-based models outperformed the BiLSTM in the presence of attractors (Figure~\ref{fig:natFig}). 
Compared to prior work with the number prediction task, our BiLSTM performed very poorly on the~4~Attractors dataset.
However, our results cannot be directly compared to previous work because of the modifications we have made to the task, data, and training procedure in order to accommodate tree-based models. 
In light of these modifications, there are several reasons why the BiLSTM's low accuracy is unsurprising. 
First, we used a balanced label distribution during training. In the standard dataset from \citet{linzen-etal-2016-assessing}, the class labels are not balanced, so models evaluated on that dataset might outperform our BiLSTM by exploiting the biased label distribution---a heuristic that our balanced training set discourages.
Another potential cause for the \mbox{BiLSTM's} poor performance is that, in order to balance the label frequencies, we used a smaller training set than was used in past work (81,000 sentences instead of 121,000 sentences).
Finally, it is possible that allowing models to see the entire sentence may allow them to acquire non-robust heuristics related to the words following the main verb. For example, a model might learn spurious correlation between the syntactic number of subjects and their direct objects.
See Appendix~\ref{sec:appFullRes}, Table~\ref{table:FullNatLangResults} for results on all test sets.

\paragraph{Constructed sentence evaluation:} With naturally occurring sentences, it is possible that models perform well not because they have mastered syntax, but rather because of statistical regularities in the data. For example, given \textit{The players *MASK* the ball}, the model may be able to exploit the fact that animate nouns tend to be subjects while inanimate nouns do not. As pointed out by \newcite{gulordava-etal-2018-colorless}, this would allow the model to correctly predict syntactic number, but for the wrong reasons. To test whether our models were leveraging this statistical heuristic, we constructed a 400-sentence test set where this heuristic cannot succeed. We did so using a probabilistic context-free grammar (PCFG) under which all words of a given part of speech are equally likely in all positions; each sentence from this grammar is of the form Subject-Verb-Object, and all noun phrases can optionally be modified by adjectives and/or prepositional phrases (see Appendix~\ref{sec:appPCFG}), as in~\ref{art1}:

\ex. The \textbf{fern} near the sad teachers \textbf{hates} the singer.\label{art1} 

\noindent
The Dependency LSTM is especially likely to fall prey to word cooccurrence heuristics, as it lacks the ability for a parent to account for the sequential position of its children. This can be an issue when determining whether a verb is supposed to be singular or plural, because the model has no robust way to distinguish a verb's subject from its direct object. 
The dependency model did indeed perform at chance (See the bar graph in Figure~\ref{fig:natFig}).\footnote{
Most sentences in the test set have only two nouns. 50\% of the time, they will agree in number, and the syntactic number is unambiguous. Random guessing on the other 50\% of cases would yield about 75\% accuracy.} This suggests that the dependency model's high accuracy is partially due to lexical heuristics rather than syntactic processing. In contrast, the other models performed well, suggesting that they are less susceptible to relying on word cooccurrence.

\section{Experiment 2: Fine-tuning}
\label{sec:exp2} 

In Experiment 1, tree-based models dramatically outperformed the BiLSTM in the presence of attractors. This difference may have arisen because most natural language sentences are simple, and thus they do not generate enough signal to illustrate the importance of tree structure to a low-bias learner, such as a BiLSTM. Recent work has shown the effectiveness of syntactically-motivated fine-tuning at increasing the robustness of neural models \cite{min_data_aug}. Would our models generalize more robustly if we added a few training examples that do not lend themselves to non-syntactic  heuristics?

To provide the model with a stronger signal about the importance of syntactic structure, we fine-tuned our models on a dataset designed to impart this signal. We used a variant of the PCFG (see Appendix~\ref{sec:appPCFG}) from Section~\ref{sec:exp1} to generate a~500-sentence \textbf{augmentation set}. This augmentation set cannot be solved using word cooccurrence statistics, and contains some sentences with attractors. The models were then fine-tuned on the augmentation set for just \textit{one epoch} over the~500 examples. See Appendix~\ref{app:exp2train} for training details.

\paragraph{Results:}
The head-lexicalized model and the BiLSTM benefited most from fine-tuning, with the head-lexicalized model now matching the performance of the Constituency LSTM, and the BiLSTM showing dramatic improvement on sentences with multiple attractors (Figure~\ref{fig:fineFig}; see Appendix~\ref{sec:appFullRes}, Table~\ref{table:FullAugResults} for detailed results). While the BiLSTM's accuracy increased on sentences with attractors, it decreased on the No Attractors test set. We suspect that this is because augmentation discouraged the model from using heuristics: while this makes performance more robust overall, it may hurt accuracy on simple examples where the heuristics give the correct answer \cite{min_data_aug}. As expected from its architectural limitations, the Dependency LSTM did not noticeably benefit from fine-tuning because it cannot extract the relevant information from the augmentation set. There was no clear effect of augmentation on the Constituency LSTM.\footnote{Note that the constructed test set used here is controlled to have no overlap with the augmentation set. Thus, it is not exactly the same as the set used in Section~\ref{sec:exp1}, but both corpora are generated from the same CFG.}

\section{Discussion}

\label{sec:discussion} Overall, we found that neural models trained on natural language achieve much more robust performance on syntactic tasks when syntax is explicitly built into the model. This suggests that the information we provided to our tree-based models is unlikely to be learned from natural language by models with only general inductive biases.

In Experiment 1, the network provided with a dependency parse did the best on most of the natural language test sets. This is unsurprising, as the task is largely about a particular dependency (i.e., the dependency between a verb and its subject). At the same time, as demonstrated by the constructed sentence test, the syntactic capabilities of the Dependency LSTM are inherently limited. Thus, it must default to non-robust heuristics in cases where the unlabeled dependency information is ambiguous. In future work, these syntactic limitations may be overcome by giving the model typed dependencies (which would distinguish between a subject-verb dependency and a verb-object dependency).

 One might expect the head-lexicalized model to perform the best, since it can leverage both syntactic formalisms.
 However, it performs no better than the constituency model when trained on natural language, suggesting that there is little benefit to incorporating dependency structure into a Constituency LSTM. In some cases, the head-lexicalized model without fine-tuning even performs worse than the Constituency LSTM. When fine-tuned on more challenging constructed examples, the head-lexicalized model performed similarly to the Constituency LSTM, suggesting that there is not enough signal in the natural language training set to teach this model what to do with the heads it has been given.
 
Our results point to two possible approaches for improving how models handle syntax. 
The first approach is to use models that have explicit mechanisms for representing syntactic structure. In particular, our results suggest that the most important aspect of syntactic structure to include is constituency structure, as constituency models appear to implicitly learn dependency structure as well. Though the models we used require parse trees to be provided, it is possible that models can learn to induce tree structure in an unsupervised or weakly-supervised manner \cite{bowman-etal-2016-fast, Choi2018LearningTC, Shen2018OrderedNI}. Another effective approach for improving the syntactic robustness of neural models is data augmentation, as demonstrated in Experiment~2. 
With this approach, it is possible to bring the syntactic performance of less-structured models closer to that of models with explicit tree structure, even with an augmentation set generated simply and easily using a PCFG.

Future work should further explore both of these approaches. Our conclusions about the importance of explicit mechanisms for representing syntactic structure can be strengthened by developing different formulations of the tree LSTMs. It seems particularly promising to explore alternative formulations of the Dependency LSTM (as mentioned above) and the effect of learning embeddings of non-terminal symbols for the Constituency LSTM. 
Finally, future work should investigate whether data augmentation can fully bridge the gap between low-bias learners and structured tree LSTMs, and whether our conclusions apply to other syntactic phenomena besides agreement.

\section*{Acknowledgments}
This research was supported by a Google Faculty Award to Tal Linzen, NSF Graduate Research Fellowship No. 1746891, and NSF Grant No. BCS-1920924. Our experiments were conducted using the Maryland Advanced Research Computing Center (MARCC).

\bibliography{acl2020}
\bibliographystyle{acl_natbib}

\clearpage
\appendix
\section{Appendix: Tree LSTM Details}
\label{sec:appEqs}

The constituency-based model that we use is the $N$-ary Tree-LSTM from \citet{tai-etal-2015-improved}, with $N$ fixed at 2 such that the tree is strictly binary; the equations for this model are shown below. Each $W$ is an input weight matrix, each $U$ is a hidden state update weight matrix, each $b$ is a bias term, each $x$ is an input word embedding, and each $h$ is a hidden state.
These equations are adaptations of the typical LSTM equations that allow the LSTM to be structured according to a constituency parse. The $x_j$ is the input embedding for a particular node in the constituency tree. In a Constituency LSTM, all leaf nodes receive the embedding for the word at that leaf, while all other nodes receive a vector of 0's. Every non-leaf node is thus a composition of the hidden states of its two children. 
In these equations, $k$ = 1 or 2, which allows ``the left hidden state in a binary tree to have either an excitatory or inhibitory effect on the forget gate of the right child'' \cite{tai-etal-2015-improved}. 
Importantly, this model distinguishes between a node's left and right children.

\begin{align}
i_j &= \sigma(W^{(i)}x_j + \sum _{l=1} ^2 U^{(i)} _l h_{jl} + b^{(i)})\label{eq:constipt}\\
f_{jk} &= \sigma(W^{(f)}x_j + \sum _{l=1} ^2 U^{(f)} _{kl} h_{jl} + b^{(f)}) \label{eq:constf}\\
o_j &= \sigma(W^{(o)}x_j + \sum _{l=1} ^2 U^{(o)} _l h_{jl} + b^{(o)})\label{eq:consto}\\
u_j &= tanh(W^{(u)}x_j + \sum _{l=1} ^2 U^{(u)} _l h_{jl} + b^{(u)})\label{eq:constu}\\
c_j &= i_j \odot u_j + \sum _{l=1} ^2 f_{jl} \odot c_{jl}\label{eq:constcell}\\
h_j &= o_j \odot tanh(c_j)\label{eq:consthidd}
\end{align}

The following equations, also from \citet{tai-etal-2015-improved}, define a child-sum Tree LSTM, which we structure according to a dependency parse. Here, the input $x_j$ is the embedding of the headword of that node in the DAG that defines a dependency parse. Note that in this model, the hidden representations of the children of a node are summed. Thus, this model cannot distinguish the linear order of its children. 

\begin{align}
\tilde{h} _j &= \sum _{k\in C(j)} h_k \label{eq:depsum}\\
i_j &= \sigma(W^{(i)}x_j +  U^{(i)}  \tilde{h}_{j} + b^{(i)})\label{eq:depipt}\\
f_{jk} &= \sigma(W^{(f)}x_j + U^{(f)}  h_{k} + b^{(f)}) \label{eq:depf}\\
o_j &= \sigma(W^{(o)}x_j +  U^{(o)} \tilde{h} _{j} + b^{(o)})\label{eq:depo}\\
u_j &= tanh(W^{(u)}x_j + U^{(u)}  \tilde{h} _{j} + b^{(u)})\label{eq:depu}\\
c_j &= i_j \odot u_j + \sum _{k \in C(j)}  f_{jk} \odot c_{k}\label{eq:depcell}\\
h_j &= o_j \odot tanh(c_j)\label{eq:dephidd}
\end{align}
\section{Appendix: Details of the Head-Lexicalized Tree LSTM Variant}
\label{sec:appHeadAlg}
Our head-lexicalized tree LSTM architecture is structured exactly the same as the Constituency LSTM. Thus, Equations~\ref{eq:constipt} through~\ref{eq:consthidd} characterize the parameters and operations performed by the head-lexicalized tree LSTM. The difference between the two architectures lies in the input, $x_j$. In the Constituency LSTM, a node $j$ was provided an input vector $x_j$ only if $j$ was a leaf node. In the head-lexicalized tree LSTM model, we use a dependency parse to generate a tag for each node in the constituency tree, which identifies which word in the corresponding constituent is the most dominant word in the dependency tree. The word embedding corresponding to the most dominant word in constituent $j$ is then provided as input $x_j$. Thus, every node in the tree receives an input vector, and the root node is guaranteed to have the headword of the whole sentence provided as input.

More formally, a dependency parse forms a tree, $T_D$. For each word, $w$, in a given sentence, denote its score, $s(w)$, as the depth of $w$ in $T_D$. A constituency parse forms a tree $T_C$. For every node $j$ in $T_C$, let $l_j$ denote the set of words corresponding to children of $j$ that are leaves of $T_C$. The input vector $x_j$ is then just the embedding of $w = \argmin _{w \in l_j} s(w)$. Ties should not exist within a constituent, but if they do (due to parsing errors), then they are broken arbitrarily.

\begin{figure}
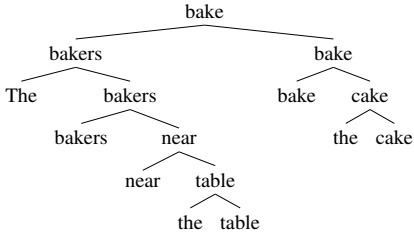

\tikzset{level distance=.8cm}
\begin{center}
\scalebox{.7}{
\Tree [ .bake [ .bakers  The [ .bakers bakers  [ .near   near  [ .table the   table ]  ] ]] [ .bake   bake   [ .cake  the   cake ] ] ] }
\end{center}
   \caption{Head-lexicalized constituency tree for the sentence \textit{The bakers near the table bake the cake}. }
     \label{fig:headConstTree}
\end{figure}

See Figure~\ref{fig:headConstTree} for an example of a head-lexicalized constituency tree. 

\section{Appendix: Training Details}
\subsection{Experiment 1} We use an embedding size and hidden cell size of 100 for every model. Our word embeddings are 100-dimensional pretrained GloVe embeddings from the Wikipedia 2014 + Gigaword 5 distribution (glove.6b.zip) \cite{penningtonglove}, and we do not tune them during training. We also employ the Adam optimizer \cite{kingma2015adam} with the PyTorch default learning rate of 0.001. Because this is a binary classification problem, we use binary cross entropy as our loss function. These hyperparameter choices are based on \citet{linzen-etal-2016-assessing}, but we increase the hidden size from 50 to 100, in order to create slightly more capacity. Though this may seem small, the models achieved high overall accuracy, suggesting that model size was not a bottleneck.

We cap training at 50 epochs, but also employ early stopping. The early stopping procedure is as follows: Train for 10,000 sentences, then evaluate on the validation data. Stop when the average decrease in validation loss over the previous five evaluations is less than 0.0005. For all models, this occurs after about 1 or 1.5 epochs. During training, the parameters that resulted in the best validation loss are saved, and these weights are used during testing. We repeat this procedure for three random initializations of each model. The reported results are averages over these three models.

In order to turn a tree LSTM into a binary classifier, we feed the hidden state of the root into a linear layer that condenses the output into a single value, and squash the result to the range [0, 1] using a sigmoid activation function. If the result of that process is greater than 0.5, then we predict label 1, else we predict label 0. For the bidirectional LSTM, we take the representation of the masked verb from both the left to right and right to left passes and feed both of these into a linear classifier. Then we repeat the process described above, using a sigmoid activation function to constrain the prediction to the range [0, 1], and classifying based on this value.
\label{app:exp1train} 

\subsection{Experiment 2}
We take the same models from Experiment 1 and fine tune them on the augmentation set. We train for one epoch with the same parameters used in Experiment 1, and then use the resulting weights to evaluate the models. \label{app:exp2train}

\section{Appendix: Data}
\label{app:data}
The original dataset contains approximately 1.3 million sentences. We use the Stanford constituency parser and Stanford dependency parser \cite{manningparser} to generate the two types of parse trees for each of these sentences, and then convert these objects into suitable representations for our models. In this process, a small percentage of examples were discarded due to the parser failing to parse them. We deviate from past work by ensuring that both classes (\textsc{SINGULAR} and \textsc{PLURAL}) are of equal size. This results in more data from the majority class (singular verb class) being thrown away. After these exclusions, we have approximately 903,000 sentences remaining. We provide our models 9\% of this (81,300 sentences) to train on, 0.1\% (904 sentences) to validate, and then generate our test sets from the remainder of the data. All sentences were stripped of quotation marks, apostrophes, parentheses and hyphens in order to minimize parsing failures.

The sizes of our test sets are as follows: No Attractors (50,000 sentences), Any Attractors (52,815 sentences), One Attractor (41,902 sentences), Two Attractors (8,473 sentences), Three Attractors (1,884 sentences), and Four Attractors (556 sentences). Note also that the Any Attractors dataset  is the union of the One, Two, Three, and Four Attractors datasets.

\section{Appendix: Full Results}
\label{sec:appFullRes}

Table~\ref{table:FullNatLangResults} contains the full results after training all models on natural language. Table~\ref{table:FullAugResults} contains the full results after augmentation.
\begin{table*}[!h]
\centering
\begin{tabular}{c c c c c} 

 \toprule
 Attractors & BiLSTM & Dependency & Constituency & Head \\ [0.5ex] 
 \midrule
 No & 96.4\% & 95.5\% & \textbf{97.3\%} & 97.2\% \\ 
 Any & 70.8\% & \textbf{91.4\%} & 90.2\%  & 87.0\%\\
 1 & 74.6\%& \textbf{91.9\%} & 91.3\% & 88.7\%\\
 2 &  59.7\% & \textbf{89.7\%} & 87.1\% &  82.0\% \\
 3 & 48.4\% & \textbf{87.6\%} & 83.7\% & 77.0\%\\ 
 4 & 41.0\% & \textbf{87.0\%} & 80.8\% & 73.1\%\\
 Constructed & 96.0\% & 73.8\% & \textbf{97.6\%} & 97.3\%\\[1ex] 
 \bottomrule
\end{tabular}
\caption{Natural language results for all datasets. Best performances are bolded. All numbers are averaged over three models.}
\label{table:FullNatLangResults}
\end{table*}

\begin{table*}[!h]
\centering
\begin{tabular}{c c c c c} 
 \toprule
 Attractors & BiLSTM & Dependency & Constituency & Head \\ [0.5ex] 
 \midrule
 No & 88.8\% & 94.9\% & \textbf{98.1\%} & 97.2\% \\ 
 Any & 77.1\% & 90.1\% & \textbf{91.9\%}  & 91.5\%\\
 1 & 76.6\%&  90.8\% & \textbf{93.2\%} & 92.8\%\\
 2 &  78.5\% & 87.9\% & \textbf{88.1\%} &  87.6\% \\
 3 & 80.1\% & \textbf{85.2\%} & 83.5\% & 83.3\%\\ 
 4 & 81.1\% & \textbf{85.9\%} & 78.4\% & 79.4\%\\
 Constructed & 95.3\% & 75.8\% & 99.7\% & \textbf{99.8\%}\\[1ex] 
 \bottomrule
\end{tabular}
\caption{Results for all datasets after augmentation. All numbers are averaged over three models.}
\label{table:FullAugResults}
\end{table*}

\section{Appendix: Probabilistic Context Free Grammars}
\label{sec:appPCFG}

Figure~\ref{fig:cfg} contains the probabilistic context free grammar used to generate the constructed corpora.
\begin{figure*}[]
\begin{center}

\newcommand{\pipe}{\kern5pt\rule[-\dp\strutbox]{.6pt}{\baselineskip}\kern5pt}

\vskip 0.12in
\fbox{\parbox{10cm}{
$\textrm{S} \rightarrow \textrm{DetP}_s \enspace \textrm{VP}_s \pipe
\textrm{DetP}_p \enspace \textrm{VP}_p$

$\textrm{DetP}_s \rightarrow \textrm{Det} \enspace \textrm{NP}_s$

$\textrm{DetP}_p \rightarrow \textrm{Det} \enspace \textrm{NP}_p$\\

$\textrm{NP}_s \rightarrow \textrm{Adj} \enspace \textrm{NP}_s \pipe \textrm{NP}_s \enspace \textrm{PP} \pipe \textrm{Noun}_s$

$\textrm{NP}_p \rightarrow \textrm{Adj} \enspace \textrm{NP}_p \pipe \textrm{NP}_p \enspace \textrm{PP} \pipe \textrm{Noun}_p$\\

$\textrm{PP} \rightarrow \textrm{Prep} \enspace \textrm{DetP}_s \pipe \textrm{Prep} \enspace \textrm{DetP}_p$\\

$\textrm{VP}_s \rightarrow \textrm{Verb}_s \enspace \textrm{DetP}_s  \pipe \textrm{Verb}_s \enspace \textrm{DetP}_p$\\
$\textrm{VP}_p \rightarrow \textrm{Verb}_p \enspace \textrm{DetP}_p  \pipe \textrm{Verb}_p \enspace \textrm{DetP}_p$\\

$\textrm{Det} \rightarrow \textrm{the}$\\

\hangindent=0.7cm
$\textrm{Noun}_s \rightarrow \textrm{plane} \pipe \textrm{plant} \pipe \textrm{bear} \pipe \textrm{bird} \pipe \textrm{car}  \pipe \textrm{dancer} \pipe \textrm{singer} \\\pipe \textrm{president} \pipe \textrm{squirrel} \pipe \textrm{cloud} \pipe \textrm{actor} \pipe \textrm{doctor} \pipe \textrm{nurse} \pipe \textrm{chair}\\ \pipe \textrm{student} \pipe \textrm{teacher} \pipe \textrm{fern} $

\hangindent=0.7cm
$\textrm{Noun}_p \rightarrow \textrm{planes} \pipe \textrm{plants} \pipe \textrm{bears} \pipe \textrm{birds} \pipe \textrm{cars}  \pipe \textrm{dancers} \\\pipe \textrm{singers} \pipe \textrm{presidents} \pipe \textrm{squirrels} \pipe \textrm{clouds} \pipe \textrm{actors} \pipe \textrm{doctors} \\\pipe \textrm{nurses} \pipe \textrm{chairs} \pipe \textrm{students} \pipe \textrm{teachers} \pipe \textrm{ferns}$\\

\hangindent=0.7cm
$\textrm{Verb}_s \rightarrow \textrm{eats} \pipe \textrm{pleases} \pipe \textrm{loves} \pipe \textrm{likes} \pipe \textrm{hates}  \pipe \textrm{destroys} \pipe \textrm{creates} \\\pipe \textrm{fights} \pipe \textrm{bites} \pipe \textrm{shoots} \pipe \textrm{arrests} \pipe \textrm{takes} \pipe \textrm{leaves} \pipe \textrm{buys} \\\pipe \textrm{brings} \pipe \textrm{carries} \pipe \textrm{kicks} $

\hangindent=0.7cm
$\textrm{Verb}_p \rightarrow \textrm{eat} \pipe \textrm{please} \pipe \textrm{love} \pipe \textrm{like} \pipe \textrm{hate}  \pipe \textrm{destroy} \pipe \textrm{create} \\\pipe \textrm{fight} \pipe \textrm{bite} \pipe \textrm{shoot} \pipe \textrm{arrest} \pipe \textrm{take} \pipe \textrm{leave} \pipe \textrm{buy} \pipe \textrm{bring} \\\pipe \textrm{carry} \pipe \textrm{kick} $\\

\hangindent=0.7cm
$\textrm{Adj} \rightarrow \textrm{fancy} \pipe \textrm{green} \pipe \textrm{handsome} \pipe \textrm{pretty} \pipe \textrm{large}  \pipe \textrm{big} \pipe \textrm{scary} \\\pipe \textrm{nice} \pipe \textrm{happy} \pipe \textrm{sad} \pipe \textrm{dangerous} \pipe \textrm{evil} \pipe \textrm{sloppy} $\\

\hangindent=0.7cm
$\textrm{Prep} \rightarrow \textrm{on} \pipe \textrm{by} \pipe \textrm{near} \pipe \textrm{around} $
}
}
\caption{Probabilistic Context-free grammar used for creating constructed datasets. For the constructed language test set, the probabilities for the three potential expansions of NP$_s$ and NP$_p$ are .1, .1, .8, respectively. For the augmentation set, these probabilities are .69, .04, .27. For all other nonterminals, all possible expansions have uniform probability in both test and augmentation sets. PPs are present in approximately one third of sentences in both the test and augmentation sets.
} \label{fig:cfg}
\end{center}
\end{figure*}

\end{document}